# Simulation of Dynamic Yaw Stability Derivatives of a Bird Using CFD

**M.A. Moelyadi\*** and **G. Sachs**§

\*Dr.-Ing., Aeronautics and Astronautics Department
Bandung Institute of Technology
Ganesha 10, 40132, Bandung, Indonesia
e-mail: moelyadi1968@hotmail.com

§Professor, Institute of Flight Mechanics and Flight Control
Technische Universität München
Boltzmannstrasse 15, 85748 Garching, Germany
e-mail: sachs@tum.de

**Abstract**

Simulation results on dynamic yaw stability derivatives of a gull bird by means of computational fluid dynamics are presented. Two different kinds of motions are used for determining the dynamic yaw stability derivatives $C_{Nr}$ and $C_{N\dot{\beta}}$. Concerning the first one, simple lateral translation and yaw rotary motions in yaw are considered. The second one consists of combined motions. To determine dynamic yaw stability derivatives of the bird, the simulation of an unsteady flow with a bird model showing a harmonic motion is performed. The unsteady flow solution for each time step is obtained by solving unsteady Euler equations based on a finite volume approach for a smaller reduced frequency. Then, an evaluation of unsteady forces and moments for one cycle is conducted using harmonic Fourier analysis. The results on the dynamic yaw stability derivatives for both simulations of the model motion show a good agreement.

## 1 Introduction

In the early days of aeronautics the flight of birds has stimulated scientists and engineers in the research and development of aircraft. Much has been learnt from nature and pioneers in the aeronautical field have studied the flight of birds, e.g. Ref. 1.

Birds move from one place to another by driving their wings in air. The motions of the bird in the air consist of flapping flight as well as gliding and soaring flight (Ref. 2). Investigations on the aerodynamic characteristics of birds have been conducted for a variety of purposes (Refs. 2 – 6). For example, an issue is formation flight of the birds as an energy saving mechanism during migration (Refs 3, 6). Other investigations deal with aerodynamic characteristics of bird wings in steady flow conditions (Refs. 4, 7). In this simulation, the bird configuration is modeled at a fixed position and the flow does not change with time. However, there are also unsteady flight conditions, like yawing and rolling motions. Such movements produce a response of the airflow at the wing changing with time. This unsteady airflow affects the lateral-directional stability characteristics of birds.

Investigations on the dynamic stability derivatives for various aerodynamic bodies using CFD have been carried out in previous research. In Ref. 8, Navier-Stokes prediction of pitch damping concerning axis-symmetric projectiles in the supersonic flight regime was performed. Computational methods were applied in Ref. 9 for determining the stability derivatives for the F16XL configuration using a low-order time dependent panel method. A reduced-frequency approach was used in Ref. 10 to determine dynamic stability derivatives at sub- and transonic Mach numbers. In Ref. 11, modelling of dynamic stability derivatives was performed for the low subsonic regime using CFD. Dynamic flight stability of a hovering bumblebee was investigated in Ref. 12, using a code based on the Navier-Stokes solution and the technique of eigenvalue and eigenvector analysis.

This paper presents results on the dynamic yaw stability derivatives of a gull wing by using the FLM in-house developed code based on Euler solution, a computer program of the Institute of Fluid Mechanics of the Technische Universität München (Refs. 13, 14). Two different kinds of model motions are applied to determine the dynamic yaw stability derivatives $C_{Nr}$ and $C_{N\dot{\beta}}$.

Concerning the first one, a lateral translation and a rotation in yaw are dealt with separately. The second one consists of a combination of translational and rotational motions. The unsteady flow solution is obtained by solving unsteady three dimensional Euler equations. The motion of the bird is modeled using harmonic functions at small reduced frequencies. The dynamic grids are generated for every time step that is adapted to the movement. Because of the motion-induced mesh deformation near the body, it is necessary to smooth the mesh using the Poisson algorithm. Then, the dynamic yaw stability derivatives can be computed applying Fourier analysis using the unsteady aerodynamic force and moment coefficients.

## 2 Dynamic Yaw Stability Derivatives

### 2.1 Motion Modellings

The forces and moments of the bird due to lateral motions can be assumed to be functions of the side slip angle and its derivative with respect to time ($\beta$, $\dot{\beta}$) as well as the rate of yaw ($r$). In coefficient form, the yawing moment, $C_N$, may be written as, using expansion series:

$$C_N = C_{N0} + \beta C_{N\beta} + \dot{\beta} C_{N\dot{\beta}} + ......$$
$$+ r C_{Nr} + \dot{r} C_{N\dot{r}} + ...... \quad (1)$$

The yaw derivatives of interest are the time rate related derivatives $C_{Nr}$ and $C_{N\dot{\beta}}$. The derivative $C_{N\dot{\beta}}$ can be obtained by moving the bird model in an appropriate motion, namely by laterally translating the model to the right and the left. For the computational simulation, a harmonic lateral motion of the gull wing is applied as shown in Fig. 1. The motion can be expressed as

$$y = y_{\max} \sin(k\tau) \quad (2)$$

$$\beta = \frac{\dot{y}}{V_\infty} = \frac{y_{\max} k \cos(k\tau)}{V_\infty} = \bar{y}_{\max} k \cos(k\tau) \quad (3)$$

$$\dot{\beta} = \frac{\ddot{y}}{V_\infty} = -\frac{y_{\max} k^2 \sin(k\tau)}{V_\infty} = -\bar{y}_{\max} k^2 \sin(k\tau) \quad (4)$$

The derivative $C_{Nr}$ cannot be determined applying a simple motion, like the lateral translation. Using an oscillation of the model about the centre gravity, the derivative combination $C_{Nr} + C_{N\dot{\beta}}$ can be obtained. In Fig. 2, a harmonic rotation about z-axis is shown. This rotation can be expressed as

$$\Psi = \Psi_{\max} \cos(k\tau) \quad (5)$$

$$r = \dot{\Psi} = -\Psi_{\max} k \sin(k\tau) \quad (6)$$

where $\Psi$ is yaw angle. The derivative $C_{Nr}$ can be determined by subtracting the derivative $C_{N\dot{\beta}}$ from the derivative combination.

Concerning the second motion modeling, the derivative $C_{Nr}$ can be directly obtained by using a combined motion, consisting of a lateral translation and a rotation in the yaw axis as shown in Fig. 3. For this modeling each motion produces a side slip effect, canceling out each other. Thus, a zero total sideslip effect results for every position.

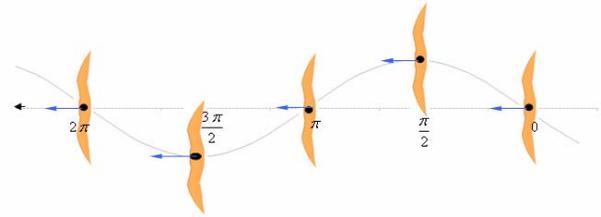

**Figure 1:** Harmonic lateral motion for derivative $C_{N\dot{\beta}}$

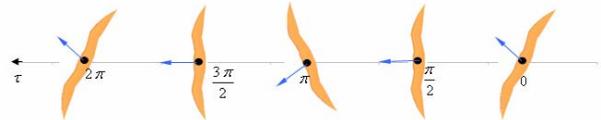

**Figure 2:** Harmonic yaw motion for derivative combination $C_{Nr} + C_{N\dot{\beta}}$

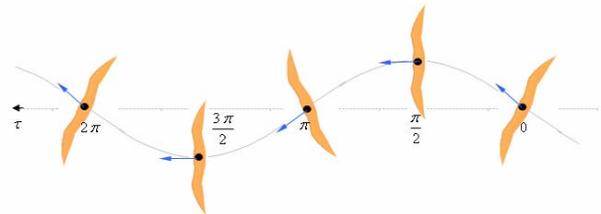

**Figure 3:** Harmonic motion consisting of a lateral translation and a yaw rotation for derivative $C_{Nr}$

### 2.2 Calculation of Lateral Stability Derivatives

The Fourier analysis is used to compute dynamic stability derivatives from the time-dependent force and moment coefficients. These force and moment coefficients are calculated by integrating the time-dependent pressure on the body surface of the moving wing. The FLM in-house developed code based on Euler solution is used to obtain time dependent flow variables such as pressure and velocity that are present both in the flow field and on the body surface. Applying $r = \dot{\Psi} = \dot{\beta}$ and $\dot{r} = \ddot{\beta}$ with regard to Eq. (1) and using an expansion series, the yawing moment coefficient $C_N$ can be written as:

$$C_N = C_{N0} + \beta(k\tau) C_{N\beta} + \dot{\beta}(k\tau) C_{N\dot{\beta}} + \ddot{\beta}(k\tau) C_{N\ddot{\beta}}$$
$$+ ...... \quad (7)$$

where $\tau$ is the time step and $k$ the reduced frequency. For a harmonic lateral motion, the following relation applies

$$\beta(k\tau) = \beta_0 + \beta_{\max} \sin(k\tau) \quad (8)$$

Thus, for the first and second derivatives

$$\dot{\beta}(k\tau) = \beta_{\max} k \cos(k\tau) \quad (9)$$

$$\ddot{\beta}(k\tau) = -\beta_{max} k^2 \sin(k\tau) \qquad (10)$$

Using Eqs. (8), (9) and (10), the following expressions are obtained from Eq. (7):

$$C_N(k\tau) = C_{v0}\beta_0 C_{N\beta} + \beta_{max}\sin(k\tau)C_{N\beta} + \beta_{max}k\cos(k\tau)C_{N\dot{\beta}} - \beta_{max}k^2\sin(k\tau)C_{N\ddot{\beta}} + \ldots \qquad (11)$$

For small amplitude and reduced frequency values, the terms of second and higher order can be neglected. Then

$$C_N(k\tau) = C_{N0} + \text{Re}[C_{N\beta}\beta_{max}e^{ik\tau} + C_{N\dot{\beta}}ik\beta_{max}e^{ik\tau}] \qquad (12)$$

The derivatives can be determined using Fourier analysis to yield

$$C_{N\beta} = \frac{\text{Re}[\overline{C}_N^1]}{\beta_{max}} = \frac{a_1}{\beta_{max}} \qquad (13)$$

$$C_{N\dot{\beta}} = \frac{\text{Im}[\overline{C}_N^1]}{k\beta_{max}} = \frac{b_1}{k\beta_{max}} \qquad (14)$$

where the coefficients $a_1$ and $b_1$ are given by

$$a_1 = \frac{1}{\pi}\int_0^{2\pi} C_N(k\tau)\sin(k\tau)d(k\tau) \qquad (15)$$

$$b_1 = \frac{1}{\pi}\int_0^{2\pi} C_N(k\tau)\cos(k\tau)d(k\tau) \qquad (16)$$

Because $C_N$ is a $2\pi$-periodic function, the determination of dynamic stability derivatives is obtained by reading off the periodically unsteady aerodynamic forces and moments given by from the Euler solution.

### 2.3 Computational Method for Determining Unsteady Aerodynamic Characteristics of a Bird Wing

For computational simulations of unsteady aerodynamic characteristics of the bird wing, the proper geometry, high quality meshes as well as a robust and accurate numerical method are required to obtain solutions of the flow field.

#### 2.3.1 Geometry of Gull Wing and Meshes

For dynamic stability derivative determination, the gull is taken as a computational model. All data used for the bird are referred to Ref. 2. The gull configuration is depicted in Fig. 4, and the wing profile in Fig. 5. The bird shows the following data:

- Area reference:    0.206 m$^2$
- Length reference:  0.160 m
- Span reference:    1.560 m
- Total length:      0.650 m
- Weight:            1.607 kg

- Minimum speed:     11.7 m/s
- Maximum speed:     20.6 m/s

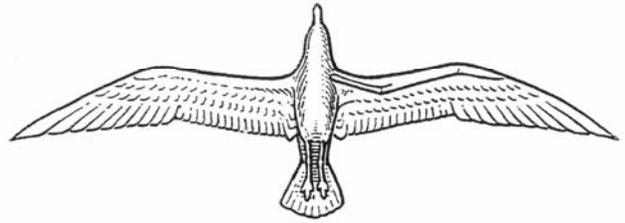

**Figure 4: Gull (from Ref. 2)**

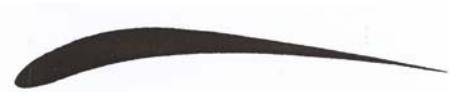

**Figure 5: Profile of gull wing**

The procedure for generating the grids for unsteady flow calculations includes block topology generation, initial mesh generation and treatment for mesh smoothing and boundary movement. The computational simulations start with the definition of the computational domain by generating a block topology. A multi-block technique is used to provide more accurate solutions rather than the use of only a single block. The proper block topology is a good start for obtaining high-quality grids and hence accurate solutions. This requires a good understanding about the details of the geometry of the model including the body shape and flow features. The generated topology for the gull wing is shown in Fig. 6. There are 60 blocks in the computational domain, namely 10 larger blocks in the farfield and 50 smaller blocks close to the wing surface. The smaller blocks are drawn to a larger scale in Fig. 7.

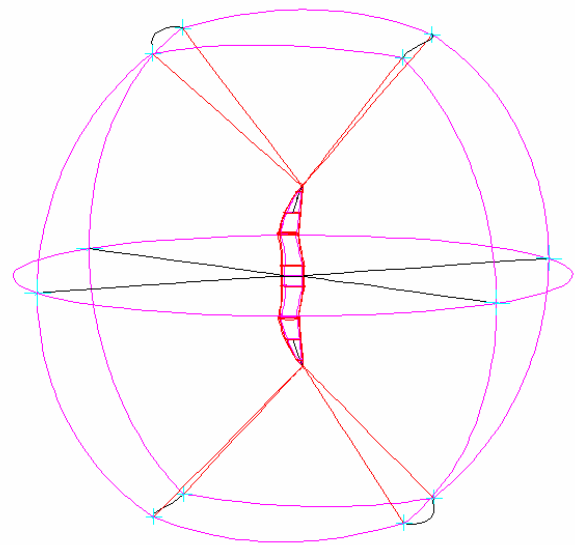

**Figure 6: Topology of gull wing**

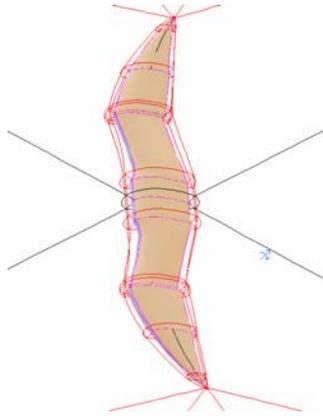

**Figure 7: Topology closed to the gull wing**

After the block topology is formed, the points for each block are distributed along the edge of the block. It is necessary to concentrate a major number of points in critical regions such as in leading- and trailing-edge sections of the wing. The point distribution becomes also denser for points closer the body surface. The initial (coarse) grids in the volume and at the surfaces of the blocks are generated by an interpolation technique called transfinite interpolation that is based on the algebraic method in the ICEMCFD software (Ref. 15). The next step is to improve the grid quality from coarse grids becoming finer grids. This is achieved iteratively by solving Poisson's equation (Ref. 16). The use of a Poisson algorithm results in smoothing the initial grid in order to yield small cell deformation and continuous cell growth. The connection between adjacent blocks is organized by mother-child relations where the grid points located at a block connection are allowed to move during the iteration process. The convergence criterion for sufficient smoothness is fulfilled if the change in the residual is below $10^{-3}$. The mesh generation for the gull wing is shown in Fig. 8.

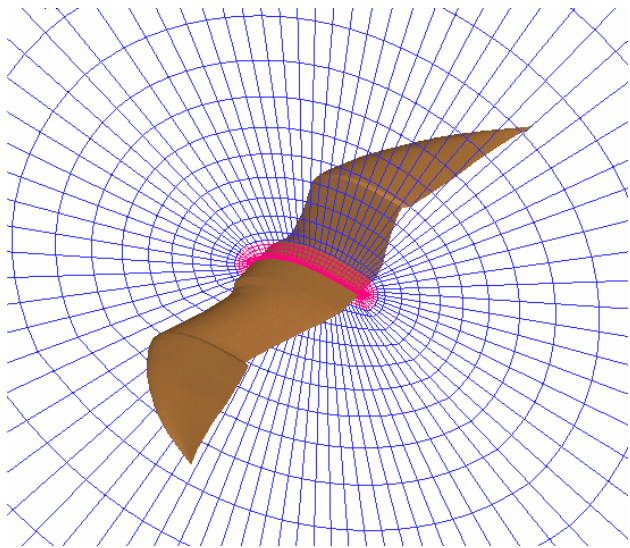

**Figure 8: Mesh generation for the gull wing**

For the simulation of unsteady aerodynamic characteristics of the gull wing, moving grids are required. The grids are adjusted and distorted by the movement of the wing surface and the fixed farfield boundary. The grids in the smaller blocks close to the wing surface move as far as the wing moves, while the grids in the larger blocks show a gradual decrease. During this deformation the quality of the computational grids has to be good at each time step in order to obtain an accurate solution. This is carried out by smoothing the grids locally using the local smoothing technique based on Laplace's solution.

For direct determination of the derivative $C_{Nr}$, the wing motion is modeled as a lateral translation and a yaw rotation. These motions can be expressed as

$$y = y_{\max} \sin(k\,\tau) \qquad (17)$$

$$\Psi = \Psi_{\max} \sin(k\,\tau) \qquad (18)$$

Results on the moving grids for simultaneous lateral and yaw motions are shown in Fig. 9.

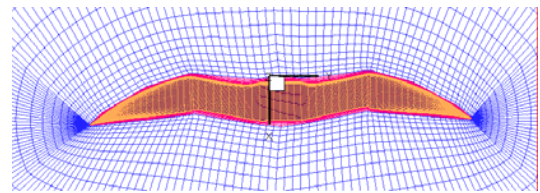

(a) y = 0.349065 chord and Ψ = 1.0 deg

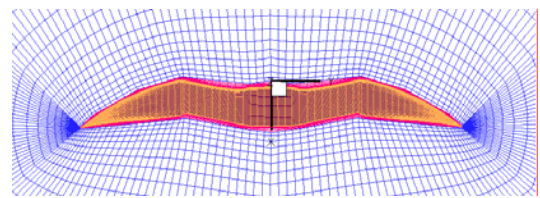

(b) y = 0.0 chord and Ψ = 0.0 deg

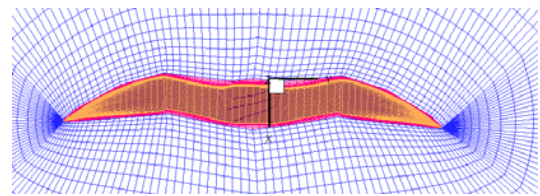

(c) y = -0.349065 chord and Ψ = -1.0 deg

**Figure 9: Dynamic grids for simultaneous lateral and lateral motions**

### 2.3.2 Numerical Flow Simulation Method

To obtain solutions in the flow field around the gull wing configuration, the unsteady Euler equations are solved using the finite volume method for spatial discretization and the first order Runge-Kutta scheme for the temporal discretization. The calculation for the flux vectors between volumes are performed using the modified AUSM method (Ref. 17). In order to guarantee the spatial second order accuracy, the left and right states at the cell interfaces are obtained with the MUSCLE approach. For unsteady flow filed calculations, a dual time-stepping methodology is used (Ref. 18). This method employs a pseudo-time to redefine the unsteady flow problem into a steady flow problem, with the physical time derivative included in the discretized equations. The conservation form of unsteady Euler equation is written in body fitted co-ordinates given by the following relation

$$\frac{\partial \mathbf{Q}}{\partial \tau} + \frac{\partial \mathbf{F}}{\partial \xi} + \frac{\partial \mathbf{G}}{\partial \eta} + \frac{\partial \mathbf{H}}{\partial \zeta} = \mathbf{0} \qquad (19)$$

where $\mathbf{Q}$ is the vector of conservative variables times the Jacobian transformation, $\mathbf{J}$, $\mathbf{F}$, $\mathbf{G}$, and $\mathbf{H}$ are the conservative fluxes with respect to the $\xi$, $\eta$ and $\zeta$ directions. The dual time-stepping approach is of second order with regard to time, yielding

$$\frac{3(\Omega)_I^{n+1} \mathbf{Q}_I^{n+1} - 4(\Omega)_I^n \mathbf{Q}_I^n + (\Omega)_I^{n-1} \mathbf{Q}_I^{n-1}}{2\Delta t} = -\mathbf{R}_I^{n+1}$$

(20)

where $\Delta t$ denotes the global physical time step. For the numerical calculation, this relation is approximated by introducing a pseudo-time variable $t^*$ and can be written as

$$\frac{d}{dt^*}\left(\Omega_I^{n+1} \mathbf{Q}_I^*\right) = -\mathbf{R}_I^*\left(\mathbf{Q}_I^*\right) \qquad (21)$$

where $\mathbf{Q}^*$ is the approximation of $\mathbf{Q}^{n+1}$. The unsteady residual including all constant terms during the time-stepping gathered is a source term given by

$$\mathbf{R}_I^*\left(\mathbf{Q}_I^*\right) = \mathbf{R}_I\left(\mathbf{Q}_I^*\right) + \frac{3}{2\Delta t}\left(\Omega_I^{n+1}\right)\mathbf{Q}_I^* - S_I^* \qquad (22)$$

$$S_I^* = \frac{2}{\Delta t}(\Omega)_I^n \mathbf{Q}_I^n - \frac{1}{2\Delta t}(\Omega)_I^{n-1} \mathbf{Q}_I^{n-1} \qquad (23)$$

To solve the pseudo problem described by Eq. (21), explicit multistage Runge-Kutta scheme is used by employing the pseudo-time step $t^*$ and the new size of the control volume $\Omega^{n+1}$. The time-marching process is started with the values of the steady solution. Then, it is continued until the values of the conservative variable at the new pseudo-time level ($\mathbf{Q}_I^*$) approximate the conservative value at $t + \Delta t$ ($\mathbf{Q}_I^{n+1}$) with sufficient accuracy. The flowchart for the calculation of unsteady gull wing characteristics is shown in Fig. 10.

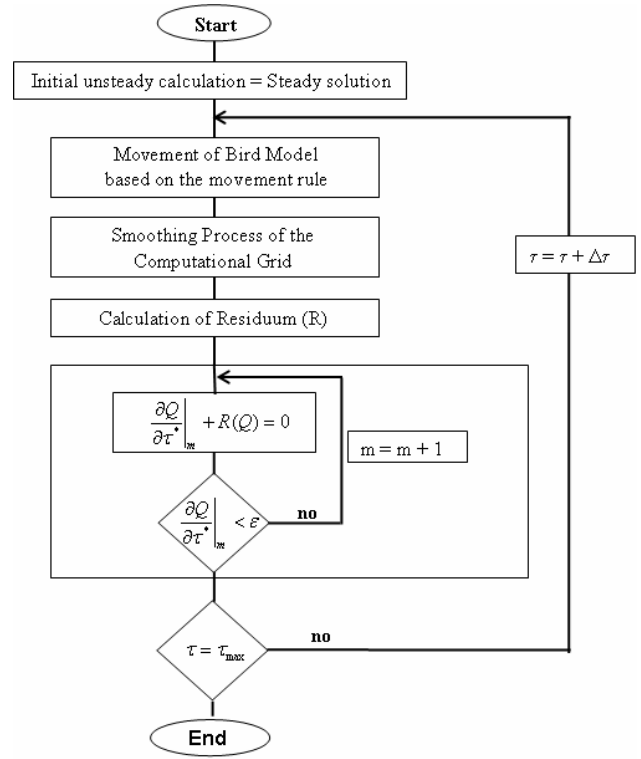

**Figure 10: Flow chart for calculation of unsteady characteristics**

## 3 Analysis of Computational Results

### 3.1 Stability Derivative Combination $C_{Nr} + C_{N\dot{\beta}}$

The solution for the derivative combination $C_{Nr} + C_{N\dot{\beta}}$ is obtained using oscillatory wing rotations in the yaw axis. In Fig. 11, the unsteady yawing moment coefficient is presented for a reduced frequency $k = 0.05$, a maximum yaw angle $\Psi_{max} = 1.0$ deg and an angle of attack $\alpha = 0$.

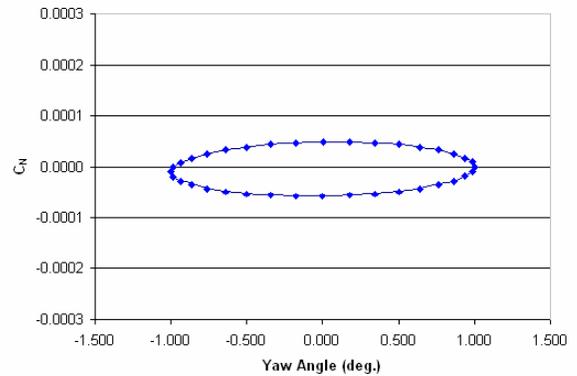

**Figure 11: Unsteady yawing moment coefficient, rotation in yaw axis**

Using the Fourier analysis in Eqs. (13) to (16), the derivative combination can be calculated to yield

$$C_{Nr} + C_{N\dot{\beta}} = 0.0612$$

### 3.2 Stability Derivatives $C_{Nr}$ and $C_{N\dot{\beta}}$

The dynamic stability derivative $C_{N\dot{\beta}}$ is calculated using the lateral translation motion model, with an angle of attack $\alpha = 0$. It is important to determine $\bar{y}_{max}$ applying $\beta = \Psi = 1.0$ deg and a reduced frequency $k = 0.05$, yielding

$$\bar{y}_{max} = \frac{\Psi}{k} = 0.349$$

In the simulation, several values are chosen for $\bar{y}_{max}$, i.e. 0.25, 0.35 and 0.45. The unsteady yawing moment coefficient applying the lateral translation motion model are depicted in Fig. 12 for these $\bar{y}_{max}$ values.

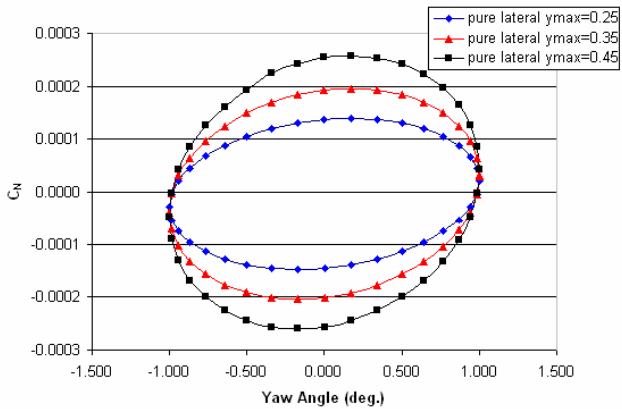

**Figure 12: Unsteady yawing moment coefficient, lateral translation**

The stability derivative $C_{N\dot{\beta}}$ is obtained using the Fourier analysis, with results as shown in Table 1. The stability derivative $C_{Nr}$ can then be obtained by subtracting the derivative of $C_{N\dot{\beta}}$ from the derivative combination, yielding results as given in Table 1.

**Table 1. Derivatives $C_{Nr}$ and $C_{N\dot{\beta}}$ for various $\bar{y}_{max}$**

| $\bar{y}_{max}$ | $C_{N\dot{\beta}}$ | $C_{Nr} + C_{N\dot{\beta}}$ | $C_{Nr}$ |
|---|---|---|---|
| 0.25 | 0.1616 | 0.0612 | -0.1004 |
| 0.35 | 0.2258 |  | -0.1646 |
| 0.45 | 0.2920 |  | -0.2308 |

This result can be confirmed considering the gradient of the side force with regard to the side slip angle. In Fig. 13, the unsteady side force coefficient depending one the yaw angle is presented, for different lateral translation values. Using the Fourier analysis, $C_{Y\beta}$ can be computed yielding the results given in Table 2.

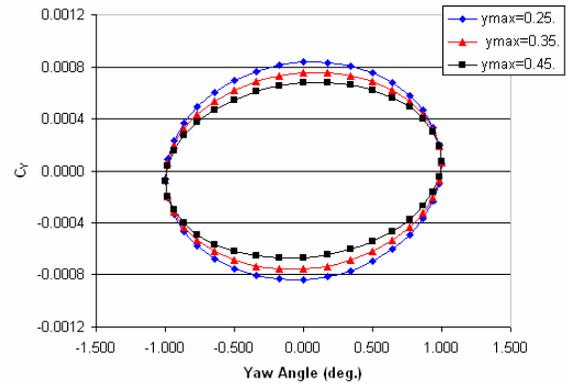

**Figure 13: Unsteady side force coefficient**

**Table 2. Derivatives $C_{Y\beta}$ and $C_{Nr}$ for various $\bar{y}_{max}$**

| $\bar{y}_{max}$ | $C_{Y\beta}$ | $C_{Nr}$ |
|---|---|---|
| 0.25 | -0.00044 | -0.1004 |
| 0.35 | 0.00009 | -0.1646 |
| 0.45 | 0.00068 | -0.2308 |

The derivatives $C_{Y\beta}$ and $C_{Nr}$ presented in Table 2 show an approximately linear dependence on $\bar{y}_{max}$. Using an interpolation technique, $C_{Nr}$ can be determined to yield

$$C_{Nr} = -0.1533$$

### 3.3 Stability Derivative $C_{Nr}$ from Combined Motion

The derivative $C_{Nr}$ can also be obtained directly using a combined motion which consists of a translation in the lateral direction and a rotation in the yaw axis (Fig. 3). The dynamic grid for the combined motion is shown in Fig. 9. The unsteady computation is performed for a reduced frequency $k = 0.05$, a maximum yaw angle $\Psi_{max} = 1.0$ deg and an angle of attack $\alpha = 0$. In Figs. 14 and 15, the yawing moment coefficient $C_n$ and the side force coefficient $C_Y$ are presented for various $\bar{y}_{max}$ values.

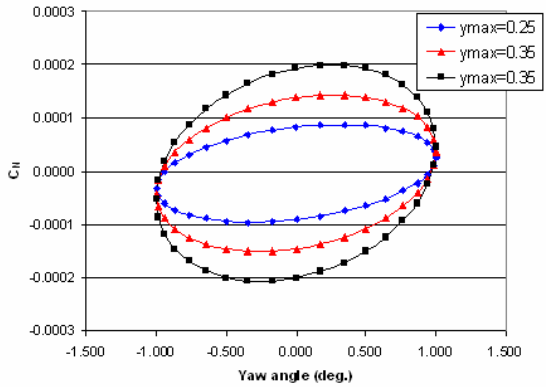

**Figure 14: Unsteady yawing moment coefficient, combined motion**

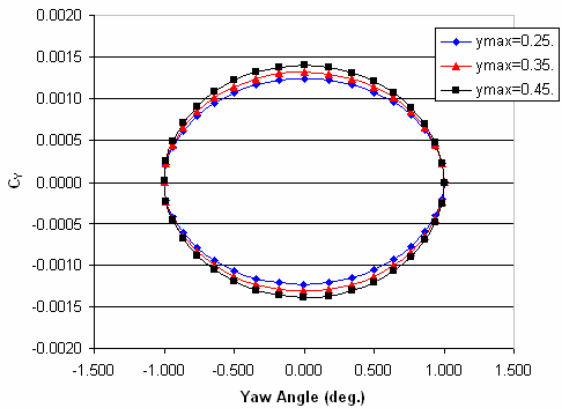

**Figure 15: Unsteady side force coefficient, combined motion**

The moment derivative $C_{Nr}$ and the force derivative $C_{Y\beta}$ are computed using the Fourier analysis. Results are presented in Table 3.

**Table 3. Derivatives $C_{Y\beta}$ and $C_{Nr}$ for combined motion**

| $\bar{y}_{max}$ | $C_{Y\beta}$ | $C_{Nr}$ |
|---|---|---|
| 0.25 | -0.00044 | -0.0998 |
| 0.35 | 0.00010 | -0.1634 |
| 0.45 | 0.00066 | -0.2265 |

The dependence of $C_{Y\beta}$ and $C_{Nr}$ on $\bar{y}_{max}$ is approximately linear. With the use of an interpolation technique, $C_{Nr}$ can be determined to yield

$$C_{Nr} = -0.1512$$

## 4 Conclusions

The determination of dynamic yaw stability derivatives of a gull wing using CFD is performed. Two different simulations of the bird model motions are used, one showing two simple harmonic motions and the other a combined motion. The simulation results provide a good agreement between both kinds of motion.